\documentclass[conference]{IEEEtran}
\IEEEoverridecommandlockouts
\usepackage{cite}
\usepackage{amsmath,amssymb,amsfonts}
\usepackage{algorithmic}
\usepackage{graphicx}
\usepackage{textcomp}
\usepackage{xcolor}
\usepackage{enumitem}

\def\BibTeX{{\rm B\kern-.05em{\sc i\kern-.025em b}\kern-.08em
    T\kern-.1667em\lower.7ex\hbox{E}\kern-.125emX}}
        
\usepackage{fancyhdr}
\usepackage[square,numbers]{natbib}
\usepackage{array}
\usepackage{booktabs}
\usepackage{url}

\usepackage{authblk}

\newcommand{\reminder}[1]{[#1]}

\newcommand{\hl}[1]{#1}
\newcommand{\camr}[1]{#1}

\newcommand{\wctan}[1]{\reminder{{\bf\small\color{cyan} (Tan)~#1}}}
\newcommand{\sara}[1]{\reminder{{\bf\small\color{green} (Sara)~#1}}}
\newcommand{\yoshi}[1]{\reminder{{\bf\small\color{blue} (Yoshi)~#1}}}
\newcommand{\behzad}[1]{\reminder{{\bf\small\color{red} (Vivian)~#1}}}
\newcommand{\alon}[1]{\reminder{{\bf\small\color{brown} (Alon)~#1}}}
\newcommand{\todo}[1]{\reminder{{\bf\small\color{red} (TODO)~#1}}}

\begin{document}

\title{Happiness Entailment:\\Automating Suggestions for Well-Being}

\author[1]{\rm Sara Evensen}
\author[1]{\rm Yoshihiko Suhara}
\author[1]{\rm Alon Halevy}
\author[1]{\rm Vivian Li}
\author[1]{\rm Wang-Chiew Tan}
\author[2]{\rm Saran Mumick}
\affil[ ]{${}^1${\it Megagon Labs}, ${}^2${\it University of Pennsylvania}}
\affil[ ]{\it {\{sara,yoshi,alon,vivian,wangchiew\}@megagon.ai, smumick@seas.upenn.edu }}


\maketitle
\thispagestyle{fancy}

\begin{abstract}
Understanding what makes people happy is a central topic in psychology. Prior work has mostly focused on developing self-reporting assessment tools for individuals and \hl{relies on} experts to analyze the periodic reported assessments. One of the goals of the analysis is \hl{to understand what actions are necessary to encourage modifications} in the behaviors of the individuals to
improve their overall well-being.

In this paper, we outline a complementary approach; on the assumption that the user journals \hl{her} happy moments as short texts, \camr{a system can analyze these texts and propose sustainable suggestions for the user that may lead to an overall improvement in her well-being. We prototype one necessary component of such a system,} the Happiness Entailment Recognition (HER) module, which takes as input a short text describing an event, a candidate suggestion, and outputs a determination about whether the suggestion \camr{is more likely to be good for this user based on the event described}. This component is implemented as a neural network model with two encoders, one  for the user input and one for the candidate actionable suggestion, with additional layers to capture \hl{psychologically significant features} in the happy moment and suggestion.
\hl{Our model achieves an AU-ROC of 0.831 and outperforms our baseline as well as the current state-of-the-art Textual Entailment model from AllenNLP by more than 48\% of improvements, confirming the uniqueness and complexity of the HER task.}

\end{abstract}


\section{Introduction}

The field of positive psychology has
made great strides in applying empirical methods to understand
happiness~\cite{Duckworth:2005:positive,Seligman:2012:Flourish,Fredrickson:1998:WhatGood,Lyubomirsky:2005:Benefits}
\camr{and find interventions that demonstrably improve it~\cite{Headey:2010:HappinessAsAChoice, Lyubomirsky:2013:SimplePositiveActivities}.}


To understand one's well-being and particularly, what actions 
lead to (un)happiness,
a wide variety of
self-reporting assessment tools have been used, such as the Day Reconstruction Method
(DRM)~\cite{Kahneman:2004:DRM}, End-of-Day Diaries (EDD), the
Experience Sampling Method (ESM)~\cite{Larson:2014:ESM} and the Ecological Momentary Assessment (EMA)~\cite{Shiffman:2008:EMS}. 
These tools are designed to capture experiences and behavior of a person as accurately a possible, and have become standard tools for measuring happiness~\cite{Diener:2000:SubjectiveWellBeing}. However, these surveys are often tedious to fill out and do not always capture the nuances of one's life.

A more recent approach to investigating happiness looks at short textual descriptions of experiences that a person had and journaled (often referred to as {\em  moments})~\cite{Jo,Asai:2018:LREC,Jaidka:2019:AffCon}. This is driven by the development of new Natural Language Understanding techniques that learn from moments that a person reports, thereby freeing the user from having to answer lengthy surveys. Some of the initial work on this approach has focused on distinguishing happy moments from unhappy ones, and finding the activities, people, and patterns involved in the reported happy moments. 

One of the main goals of positive psychology is to discover behaviors that contribute to an individual's happiness in a sustainable 
fashion.
\hl{A main barrier to this approach is the tendency for people to mis-estimate how happy things will make them~\cite{Lyubomirsky:2008:TheHowOfHappiness}. Most people spend significant time and resources on activities with the goal of improving their well-being, only to find that the affect boost they gain is smaller or shorter lasting than they had hoped. \camr{Hindsight, especially from an outside observer, is a  more reliable indicator of which activities actually foster happiness. }

Making individually-tailored suggestions has long been in the domain of psychology professionals, but \camr{we \hl{propose} a design for a system} that can supplement professional help by generating suggestions for well-being, \camr{especially in one of the numerous places where there is a shortage of mental health professionals \cite{albee1959mental}}. This system would (1) discover sustainable suggestions for activities, and (2) identify which suggestions make sense for individual users. As an input to this system, users create short journal entry ``happy moments,'' an intervention proven to improve subjective well-being~\cite{Pennebaker:1986:ExpressiveWriting}. 
The relationship between these two tasks is \camr{shown} in Fig. \ref{fig:system_overview}.

\begin{figure*}[h]
    \centering
    \includegraphics[width=0.99\textwidth]{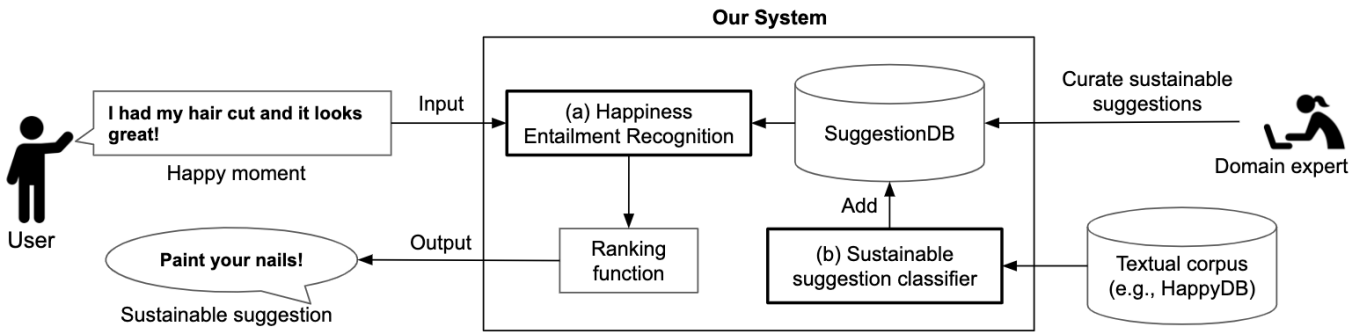}
    \caption{Overview of our system. The system outputs a best followup suggestion based on an input happy moment provided by the user. The suggestion candidates can be either manually crafted or automatically filtered from an external corpus such as HappyDB.}
    \label{fig:system_overview}
    \vspace{-1.5em}
\end{figure*}

For the first task, we define suggestions as {\em sustainable} in the sense they improve short-term well-being without being detrimental in the long term, even if the suggestion is repeated. For example, a user may have reported that adopting a puppy or buying a new car has made them happy. However, we would not want to recommend doing them again because the repetition is not sustainable. In contrast, activities like going for a swim or a walk in the park have the potential to improve the well-being of the user and are repeatable, so we consider them sustainable suggestions. Such suggestions become part of our SuggestionDB. Ideally, domain experts (e.g., psychologists) would create those sustainable suggestions, but we also develop a classifier that \camr{identifies happy moments with sustainable activities}, and confirm that the classifier achieved an AU-ROC of 0.900.

For the second of these tasks, we introduce the {\em happiness entailment recognition problem} (HER), inspired by the problem of recognizing textual entailment (RTE)~\cite{dagan2005pascal} in NLP: given a set of happy moments reported by a user, and a set of possible suggested activities, which of the activities is likely to make the user happy?  
We develop a neural network \hl{(NN)} model enhanced with information about \hl{concepts}, agency, and sociality to predict with \hl{an AU-ROC of 0.831} whether a suggestion fits a happy moment. 
\hl{``Concepts'' represents frequently occurring topics in HappyDB as a one-hot vector. ``Agency'' and ``sociality'' are binary indicators of whether the author had control over their happy moment, and whether there were other people involved, respectively. We claim that these classifications of the texts capture psychological perspectives that are important for the task.
}
}

Experimental results show that for this task, the proposed method outperforms the state-of-the-art textual entailment
recognition model, as well as 
our baseline. 
The results support that
the HER task is significantly different from conventional textual
entailment task, and that adding psychological perspectives to the model improves the performance.

Our contributions are as follows:
\begin{itemize}
    \item We describe a system for making sustainable suggestions to a user based on her description of a positive event that happened to her.
    
    \item We develop a sustainable suggestion classifier that filters
    ``suggestible'' happy moments from a corpus, and show that it performs robustly.

    \item We describe our approach to obtaining sustainable suggestions and 
    our solution for the HER problem:
    a neural network-based method that incorporates psychological features.
    Our HER model 
    \hl{outperforms both our baseline and the
    state-of-the-art textual entailment recognition model.} 


\end{itemize}

\begin{figure*}
    \centering
    \scriptsize
    \begin{tabular}{|p{6.5cm}|p{5.0cm}|p{5.0cm}|}\hline
Paint your nails & Take a hot bath or shower & Go for a walk \\\hline
Spend time with your significant other & Cuddle with your pet & Plan a vacation \\\hline
Write a letter to someone you miss & Bake cookies & Try making a new recipe \\\hline
Invite someone to lunch & Meditate & Make a to-do list for tomorrow \\\hline
Go for a bike ride & Find a restaurant you'd like to try this week & Stretch \\\hline
Be a tourist in your own town and go sightseeing. & Spend some time outdoors. & Read a book \\\hline
Be affectionate towards a loved one. & Make a new playlist for yourself & Listen to music \\\hline
Write down three things that made you happy in the last 24 hours. & Make a playlist for someone & Exercise your pet \\\hline
Be a tourist in your own town and go sightseeing. & Clean your desk or work space & Draw or paint something \\\hline
Take a photo of something that makes you happy & Make your bed & Plan a get-together \\\hline
Watch your favorite TV show & Write a thank-you card & Write a poem \\\hline
Go to a farmers market & Watch the sunset & Go swimming \\\hline
    \end{tabular}
    \caption{Manually crafted 36 sustainable suggestions.} 
    \label{fig:suggestions}        
    \normalsize
    \vspace{-1.5em}    
\end{figure*}

\section{The Happiness Entailment Recognition Task}\label{sec:HER}
Our goal is to identify which activity suggestions help improve the user's well-being and are sustainable. Such suggestions are essential 
for building a system to help the user improve her well-being such as Jo~\cite{Jo}. 

Ideally suggestions \camr{would} be based on a collection of happy moments from the user, knowledge about the user, and knowledge about the user's surroundings such as location and weather. In this paper, we explore the simplest formulation of this problem 
\camr{where we examine
only one happy moment to determine a potential suggestion that would make a user happy.} 

For example, ``paint your nails'' may be a good sustainable suggestion to a person who describes a happy moment about \hl{a} haircut, indicating that personal grooming boosts their affect.
The suggestion is less relevant if the happy moment describes ``painting the wall.'' Therefore, it requires common sense and inference to understand the relationship between a happy moment and a sustainable suggestion and to evaluate if the suggestion would make the author happy. 

\hl{We formulate this} as a binary classification task, where the system takes two input sentences, namely a happy moment and a suggestion, and classifies the pair into two classes: (1) ``entailment'' or (2) ``non-entailment,'' following the original problem formulation of the RTE task \cite{dagan2005pascal}. Compared to RTE, which has mostly focused on logical reasoning, HER considers a more subjective version of entailment based on implicit cues in a happy moment.


Because the ground truth labels for HER are given by human annotators, our model can learn  common sense about the relationship between event(s) that made the author happy and a sustainable suggestion. Averaging the results of non-expert annotations has very high agreement with expert labels for NLP tasks~\cite{snow2008cheap}, so we use crowdsourcing for human annotations.
We hope that this task can lead to insights into the causes of happiness, since predicting entailed suggestions requires an understanding of the (possibly implicit) reasons that a moment made the user happy.
%

\section{Obtaining Training Data for HER}\label{sec:corpus}
In this section, we describe the dataset creation methodology for the HER task.
As we described in Section \ref{sec:HER}, each sample consists of (1) a happy moment, (2) a sustainable suggestion, and (3) a label for the pair. We filtered happy moments from HappyDB and manually crafted sustainable suggestions before generating pairs based on them for crowdsourced annotations.

\noindent
{\bf Sustainable Suggestions~} We manually crafted 36 sustainable suggestions based on the happy moments in HappyDB. We filtered the happy moments written in a single sentence, and then created clusters based on the verbs. This helped us come up with different types of activities that should have reasonable coverage with respect to the happy moments in HappyDB.  The curated sustainable suggestions are shown in Fig. \ref{fig:suggestions}.
Note that our system can take other sustainable suggestions as input. In Section \ref{sec:automatic_suggestion}, we will describe how to expand this set by filtering sustainable suggestions from happy moments.

\noindent
{\bf Label Collection~} 
We used Amazon Mechanical Turk (MTurk)
for crowdsourced annotations. \hl{We recruited only MTurk Master Workers
to ensure high-quality annotations.}
Each task consists of $N$ (= 3 or 5) happy moments reported by different people and 1 sustainable suggestion randomly chosen from 30 in our manually curated set. An example task is shown in Fig.~\ref{fig:instructions}. For each task, 5 workers were asked to choose the happy moments for which they considered the given sustainable suggestion would be appropriate\footnote{\camr{With limited information about the user, we cannot be certain that the suggestion is a good fit, but we can leverage the common sense of annotators to infer that conditional on the happy moment, it is more likely to be a good fit than previously expected.}}.
%
%
Lastly, we left an optional free-response text box with the question about the reason and evidence of the annotation.
%
This served two purposes: including such a field for explanations has been shown to improve the quality of annotations, 
and it helped us understand cases where annotators disagreed. \hl{121 unique workers completed these tasks. }

We were concerned that the neighboring happy moments that a happy moment is presented with (the ``context'') might bias the annotations assigned. 
To avoid this effect, we made sure that happy moments were shuffled and presented with different contexts for different workers. We also experimented with showing workers either $N=3$ or $5$ happy moments at a time \hl{but found no significant difference}.

\begin{figure}[!t]
\scriptsize
\fbox{
\parbox{0.468\textwidth}{
\underline{\bf Instructions for workers ($I_1$):}\\
\noindent
Three people reported the following happy moments:
\begin{itemize}[leftmargin=*]
\vspace{-0mm}
\item[] \hspace{-1em} $\square$ I met with my city mayor and given valuable advice for development of city.
\vspace{-0mm}
\item[] \hspace{-1em} $\square$
I went on a successful date with someone I felt sympathy and connection with.
\vspace{-0mm}
\item[] \hspace{-1em} $\square$
The cake I made today came out amazing. It tasted amazing as well.
\vspace{-0mm}
\end{itemize}

Which (if any) of these people should we make the following suggestion? Check the appropriate boxes.
\begin{itemize}[leftmargin=*]
\vspace{-0mm}
\item[] Try making a new recipe
\end{itemize}
Why? There should be evidence in the happy moments for you to make this suggestion.
}}
\normalsize
\caption{Instructions $I_1$ with positive examples}
\label{fig:instructions}
\vspace{-1em}
\end{figure}

\noindent
{\bf Confidence Filtering~}
We obtained 5,387 labels (entailment or non-entailment) that were computed by taking the majority vote of the 5 annotators for the happy moment and sustainable suggestion pairs.
To ensure quality, we applied confident filtering to the annotations. For the positive (entailment) class, we filtered only rows where at least four of the five annotators agreed. For the negative (non-entailment) class, we applied more rigorous agreement standards \hl{to minimize the effects of default bias}, using only those which \hl{had a} unanimous agreement between annotators. 
Table ~\ref{table:examples} shows examples of pairs of happy moments and sustainable suggestions, including their original annotations.
The negative examples still outweighed the positive examples, so we balanced the dataset by randomly sampling a subset of the negative examples.
This filtering \hl{left} 1,364 examples, which were split into 1,068 training samples, \hl{and} 148 samples each for validation and test. 
Basic statistics of the final dataset after the filtering are shown in Table \ref{table:dataset_stats}.

\begin{figure}[t]
    \centering
    \includegraphics[width=0.22\textwidth]{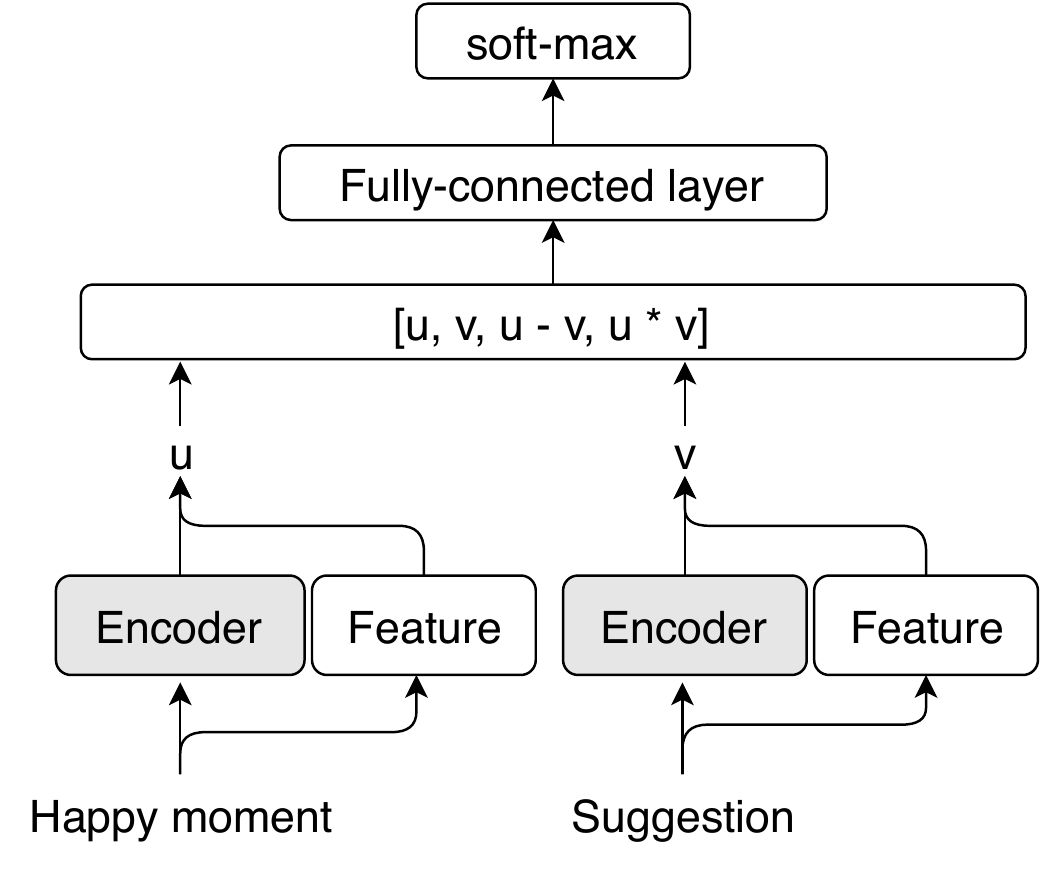}
    \vspace{-1em}
    \caption{Network architecture of the model. The encoder (in gray boxes) shares the parameters and converts input sentences (i.e., a happy moment and a sustainable suggestion) into embedding vectors. Additional feature(s) (shown as ``Feature'') that encode each input into a vector representation can be incorporated to the model. The encoded vectors will be merged into $u$ and $v$ respectively and concatenated as $[u, v, u-v, u*v]$, \hl{which is a common technique to take into account combined information of $u$ and $v$~\cite{Conneau:2017:EMNLP}}, before feeding it into the fully-connected layer and the output layer.}\label{fig:dual_encoder}
    \vspace{-1.5em}
\end{figure}

\section{HER model}\label{sec:her_model}
In this section, we describe \hl{our proposed model} for the HER task.  Our basic HER module is implemented as a dual-encoder (DE) model~\cite{Conneau:2017:EMNLP}, which is a standard NN model for NLI tasks. As shown in Fig.~\ref{fig:dual_encoder}, it has a shared encoder for both inputs (i.e., a happy moment and a sustainable suggestion.) 
Note that a model for the HER task has to be able to take arbitrary two textual inputs to judge the entailment, and thus cannot define the classes (i.e., multi-class classification) in advance. 

\noindent
{\bf Psychological Features~} As variants of the DE model, we enhanced the input happy moments and sustainable suggestions with additional \hl{psychological} features: (1) concept, (2) agency, and (3) sociality. Fig.~\ref{fig:dual_encoder} shows the architecture where the ``Feature'' box represents the additional features (i.e. concept, agency, or sociality).

We hypothesized that evidence for a sustainable suggestion often comes from overlapping concepts in the happy moment and the sustainable suggestion. For example, if the happy moment and sustainable suggestion both mention exercise, that may be evidence for happiness entailment. This hypothesis \hl{was} motivated by the concept of {\it learning generalization} in psychology---a person responds similarly to similar stimuli~\cite{Gluck:2011:LearningAndMemory}. We approximate similarity by using 15 concepts which were developed for use in the Cl-Aff Happiness Shared task. The concepts were chosen from a series of experimental MTurk tasks that asked workers to write a noun describing the event in a happy moment. The 15 most frequently occurring responses were chosen as concepts.
Table~\ref{table:concepts} shows these 15 concepts and their coverage in HappyDB.
We use Logistic Regression classifiers using bag-of-words features. 
The combined classifiers output a 15 dimensional ``concept vector,'' where each index represents the presence or absence of a concept (with a 0 or a 1). The concept vectors for happy moments and sustainable suggestions share a fully-connected layer as an encoder, whose output is concatenated to the embedding of the happy moment or suggestion. 

%

%
Similarly, we \hl{use} two Logistic Regression classifiers on bag-of-words features to predict agency and sociality labels for happy moments, \hl{which are concatenated into embedding vectors in the same manner as concept vectors.}
The {\it sociality} label captures whether a text involves people other than the author. Because humans are highly motivated to develop and maintain strong relationships \cite{needtobelong}, and cultivation of social relationships is linked with happiness \cite{Lyubomirsky:2008:TheHowOfHappiness}, we hypothesize\hl{d} that sociality labels could encourage the model to learn when social interaction was critical to a happy moment, thus should be suggested in an activity. 
The {\it agency} label describes whether the author is in control or not. Exercising agency in the form of pursuing autonomously chosen personal goals is shown to improve well-being \cite{cantor200312}, so this feature may similarly help the model learn how agency plays a role in what should be suggested. By nature, all the suggestions are agentic since the user must have the ability to engage in this activity.
%

By providing these psychological features, we allow the model to learn which features generalize between happy moments and sustainable suggestions, and possibly about psychological concepts that generalize to each other in the context of HER.
%


%

\begin{figure*}
    \centering
    \scriptsize
    \begin{tabular}{|l|p{11cm}ll|}\hline
     & Happy moment & Annotation & Sustainable suggestion \\\hline\hline
    1 & I was able to find time to go have my hair cut, something I have been putting off all month and it looks great!  & \begin{tabular}{l} {\bf entailment} \\ T T T T T \end{tabular} & paint your nails \\\hline
    2 & I met my childhood friend after a long time and got a hug from her and that moment was wonderful. & \begin{tabular}{l} {\bf non-entailment} \\ F F F F F \end{tabular} & Exercise your pet \\\hline
    3 & I was able to go and workout with a coworker of mine and that prepped me for the day and made me happy. & \begin{tabular}{l} {\bf entailment} \\ T T T T F \end{tabular} & Listen to music \\\hline
    4 & I received a call letter from the University for my job & \begin{tabular}{l} {\bf uncertain} \\ T T F F F\end{tabular} & Clean your desk or work space \\ \hline
    5 & I got to finish a tv series I was watching with my mom. & \begin{tabular}{l} {\bf uncertain} \\ T F F F F\end{tabular} & Make a to-do list for tomorrow. \\ \hline
    \end{tabular}
    \caption{Randomly chosen examples from our dataset, shown with both the selected gold labels (in bold) and the full set of annotations from the individual workers. ``unclear'' labels mean that the row was excluded from the training set.}
    \label{table:examples}
    \normalsize
    \vspace{-1.5em}
\end{figure*}

\begin{table}[t]
\centering
\caption{Statistics of the training dataset.}\label{table:dataset_stats}
\begin{tabular}{|c||c|}\hline
\# of samples (\# of pos / \# of neg) & 1364 (682 / 682) \\\hline
\# of distinct happy moments (HM) & 709 \\\hline
\# of distinct sustainable suggestions (S) & 30\\\hline
\% agency (HM, S) & 78.3, 100 \\\hline
\% sociality (HM, S) & 45.2, 9.5 \\\hline
\end{tabular}
\end{table}

\section{Evaluation}\label{sec:evaluation}
In this section, we describe our baseline models and evaluation method. We selected two baseline methods: a concept-based classifier and a textual entailment classifier. We compared these to our HER models, described in section \ref{sec:her_model}.

\noindent
{\bf Concepts Model~} Our baseline model uses the concept classifiers to detect concepts for the happy moment and the sustainable suggestion, and outputs a positive label for happiness entailment if they share any in common. 
\hl{We trained these classifiers based on HappyDB, which achieved F-scores between 0.67 and 0.90.}

\begin{table}[!t]
\centering
\caption{Most Common Concepts in HappyDB.}
\scriptsize
\begin{tabular}{|p{1.25cm}|>{\raggedleft\arraybackslash}p{0.85cm}|p{1.2cm}|>{\raggedleft\arraybackslash}p{0.85cm}|p{1.25cm}|>{\raggedleft\arraybackslash}p{0.85cm}|}\hline
    Concept & HM (\%) & Concept & HM (\%) & Concept & HM (\%) \\ \hline\hline
     Family      & 23.04 &     Romance      & 5.71   &      Technology      & 3.25 \\
     Food      & 13.44 &       Conversation & 4.89  & Weather      & 2.61\\
     Entertainment  & 12.08 & Exercise      & 4.57  &      Party      & 2.54 \\
     Career      & 10.06 &        Education      & 4.32  &      Vacation      & 2.44 \\ 
     Shopping      & 6.29 &        Animals      & 4.17  &      Religion      & 1.01 \\ \hline
     \multicolumn{4}{c|}{} & \bf{All} & \bf{78.15} \\\cline{5-6}
\end{tabular}
\normalsize
\label{table:concepts}
\vspace{-1.5em}
\end{table}

\noindent
{\bf Textual Entailment Model~} As we mentioned, the RTE task can be formulated as the same classification problem as the HER task. The task takes two different sentences and outputs a single output from the entailment, neutral, or contradiction classes. To verify the overlap between the RTE and HER tasks, we evaluated a pre-trained textual entailment recognition model. Specifically, we used the AllenNLP textual entailment (TE) model~\cite{Parikh:2016:AllenNLPTE} that is trained on the Stanford NLI (SNLI) dataset~\cite{snli:emnlp2015} since it is considered one of the state-of-the-art methods and a de-facto standard baseline method for the RTE task. We found that changing the suggestions to a first-person conjugation (e.g., ``Go for a walk'' becomes ``I should go for a walk'') improved the performance of the TE model, so we applied this as a pre-processing step to all the suggestions for this model only.

\noindent
{\bf DE Model~}
The DE model was implemented using PyTorch. The word embeddings were initialized with 300-dimension pre-trained GloVe embeddings, using a vocabulary size of 1,643. The model used a shared encoder that is a 3-layer bidirectional LSTM with 500 hidden states. 
We used Adam optimization with an 
initial learning rate of $8.28643 \times 10^{-4}$.
We used a dropout rate of $5.87755 \times 10^{-1}$,
a batch size of 32 and 
trained for 30 epochs. 
These parameters were chosen \camr{on the validation set} by training 9 versions of the model with different hyper-parameters using randomized 
search~\cite{Bergstra:2012:RandomSearch}.

\noindent
{\bf DE + Feature(s) model~} 
\hl{We used the same concepts model as the baseline model.}
The agency and sociality models were trained on the dataset of 10k labeled happy moments presented at the CL-Aff Happiness Shared Task~\cite{Jaidka:2019:AffCon}, and they achieved accuracy values of 0.898 and 0.950 \hl{for the agency and sociality classification tasks respectively}\footnote{These models were not chosen as baselines because they achieved accuracy the same as for random guessing (0.51--0.53 AU-ROC).}.
%
%
In addition to testing these three features separately, \hl{we trained and evaluated a DE model enhanced with all three features.} 
%

%

\noindent
{\bf Experimental Settings~}
For fair comparison\hl{s}, we used the same dataset split into training, validation, and test data as described in Section~\ref{sec:corpus}.  The dual-encoder methods were trained on the training set and tuned using the validation set, and all the methods were evaluated on the test set. 
%
%
\camr{
We used AU-ROC as an evaluation metric as it is considered the standard evaluation metric for binary classification. AU-ROC measures the probability that a randomly chosen positive test sample will be scored higher than a randomly chosen negative test sample, and the AU-ROC value for random guessing will be 0.5.
}

\subsection{Results and Discussion}
Table~\ref{table:results} shows the AU-ROC values on our test set of the 7 models described above.  The TE model, although state-of-the-art for the RTE task, performs about the same as random guessing for the HER task. This shows that the HER task is significantly different from the conventional RTE task, and we need a HER corpus, not an RTE corpus, to train a model. 
Table~\ref{table:results} also shows that the concept model performs only slightly better than random guessing, mostly due to false negatives. For example, the happy moment ``Coffee has made me incredibly happy'' and the sustainable suggestion ``meditate'' was labeled as entailment, but they share no concept from our list in Table \ref{table:concepts}, \camr{although they both reflect mindfulness.} Our dual-encoder model is able to predict this relation, both with and without the added features.

\camr{
Based on the performance, sociality is the most informative feature when combined with the dual-encoder model, for this task. This indicates that the sociality label in many cases influences what type of sustainable suggestion is well-suited. This is especially interesting because there is no correlation between the sociality label (of either the happy moment or the suggestion) and the entailment label, so the labels alone do not predict whether the sustainable suggestion matches the happy moment.  Rather, the model seems to learn for which happy moments social interaction is essential to the cause of happiness, and thus should be part of a sustainable suggestion. 

The best performing model includes all three additional features, but it is not clear that the difference in performance is statistically significant.
\hl{Future work includes further investigation to determine why the features interact this way, testing the importance of social interaction in suggestions, and discovering other informative psychological features of texts.} }

\begin{table}[!t]
\centering
\caption{Experimental results for the HER task.}
\label{table:results}
\begin{tabular}{|l|c|l|c|}\hline
Method   & AU-ROC & Method & AU-ROC \\\hline\hline
Concept Model &  0.561 & Dual-encoder & 0.770  \\\hline
AllenNLP Entailment & 0.548 & ~~+ concept (C) & 0.748 \\\hline
\multicolumn{2}{c|}{} & ~~+ agency (A) & 0.748 \\\cline{3-4}
\multicolumn{2}{c|}{} & ~~+ sociality (S) & 0.819 \\\cline{3-4}
\multicolumn{2}{c|}{} & ~~+ A + S & 0.806 \\\cline{3-4}
\multicolumn{2}{c|}{} & ~~+ C + A + S & {\bf 0831} \\\cline{3-4}
\end{tabular}
\vspace{-2em}
\end{table}

\section{Sustainable Suggestion Classification}\label{sec:automatic_suggestion}
The sustainable suggestions we used in our HER task were manually curated based on ideas from happy moments and positive psychology \cite{Lyubomirsky:2008:TheHowOfHappiness}. We attempt to reduce this bottleneck by building a classifier that automatically filters candidate suggestions from a corpus of happy moments such as HappyDB.

We hypothesize that happy moments contain many sustainable activities that we can suggest if we learn how to extract them. Therefore, we build a model that takes a happy moment as input and determine\hl{s} if the activity is suggestible. We define two criteria for {\it suggestibility}: (1) the activity can be repeated with a reasonable amount of effort ({\it repeatability}), and (2) it would likely make someone happy to repeat this ({\it sustainability}).

Examples of happy moments that are {\it repeatable} but not {\it sustainable} are ``I adopted a puppy'' and ``I bought a new air conditioner.'' There is a limit to how many puppies a user should adopt, and buying a new air-conditioner is probably will only make one happy if the air-conditioner is needed.

At first we attempted to crowdsource sustainable suggestions by giving MTurk workers the prompt ``Imagine a friend of yours is bored, or in a bad mood. What are some things you might suggest they do to cheer up?'', followed by three boxes for free text entries. Most of the results were either not repeatable or not sustainable, so we explore an automated way of extracting sustainable suggestions.

\subsection{Methods}
\noindent
{\bf Label collection~}
Because the notion of \hl{suggestibility} only applies to single activities, not entire happy moments, we selected only happy moments with a single activity to label\footnote{The number of activities was defined as the number of consecutive sequences of conjugated or modal verbs, as detected by the constituency parsing model of AllenNLP~\cite{Stern:2017:AllenNLPConstituencyParser}.}.
For each selected happy moment, 5 workers on MTurk answered the questions ``Is it in the author’s control to repeat this activity?'' and ``Do you think it would make the author happy to repeat this activity (assuming they could)?'' to elicit labels for our two criteria (i.e., repeatability and sustainability.) Happy moments with 4:1 agreement for both of these labels were added to the training data. Although both of our criteria are subjective, we found that filtering by agreement this way yielded reasonable results. 

For each happy moment, we labeled it as suggestible if it was rated as repeatable and sustainable by at least 4 of 5 annotators. It was labeled as not suggestible if it was rated ``no'' for repeatably or for sustainability by at least 4 out of 5 annotators. \hl{Examples are shown in Fig.~\ref{fig:labeled_suggestions}, and Table~\ref{table:suggest_data_stats} shows more details about the dataset.}

\begin{figure}
\centering
\scriptsize
\begin{tabular}{|p{4cm}||p{4cm}|}\hline
     Suggestible &  Not Suggestible \\ \hline\hline
      Today I used the KonMari method to clean my home. & I finished my first year in college with good marks\\\hline
      Taking a bubble bath. & I got to go to my cousin's wedding. \\\hline
     I watched Doctor Strange. & I got accepted into a doctoral program. \\\hline
\end{tabular}  
\caption{Examples of happy moments and their suggestibility labels.
}\label{fig:labeled_suggestions}
\normalsize
\vspace{-1em}
\end{figure}

\begin{table}[!t]
\centering
\caption{Dataset statistics for suggestibility prediction. }
\label{table:suggest_data_stats}
\begin{tabular}{|c||c|}\hline
\# of samples (\# of pos / \# of neg) & 999 (440 / 559) \\\hline
\# of filtered samples (\# of pos / \# of neg) & 290 (150 / 140) \\\hline
\end{tabular}
\vspace{-2em}
\end{table}

\noindent
{\bf Experimental Setting~}
For the suggestibility prediction task, we first trained a simple Logistic Regression model trained on Bag-of-Words features. We also implemented a Bi-LSTM RNN model with self-attention (Bi-LSTM)~\cite{Lin:2017:SelfAttention}, which is a standard technique for text classification tasks in NLP.
%
We split the data into 70\% training, 15\% validation, and 15\% test sets, \hl{and trained and evaluated both models on the training and test sets respectively.}

\subsection{Results and Discussion}
The results are shown in Table \ref{table:suggest_results}. Both the Logistic Regression model and the Bi-LSTM model performed well, with the Bi-LSTM performing better. This indicates that our sustainable suggestion classification framework, regardless of machine learning algorithms, accurately filters sustainable suggestion candidates from a corpus of happy moments.

\begin{table}[t]
\centering
\caption{Experimental results for the suggestibility prediction task.}\label{table:suggest_results}
\begin{tabular}{|l||c|}\hline
Method & AU-ROC \\\hline\hline
Guessing all positive & 0.500  \\\hline
Logistic Regression & \hl{0.867} \\\hline
Bi-LSTM & {\bf 0.900} \\\hline
\end{tabular}
\vspace{-2em}
\end{table}

The bag-of-words features allow us to see which words carry the most weight in the classification decision of the Logistic Regression model. The model has highest weights for words that relate to spending or winning, as this is a strong indicator that an event is not suggestible (winning is usually out of the author's control, and spending money is usually not labeled as sustainable.) Words relating to family members or friends also had high weights, as most social activities are sustainable and repeatable. This simple model works quite well for our task because words like this carry a lot of information about whether the activity is suggestible.


We consider two major limitations of this method as follows.
First, many happy moments are similar, so we need post-processing to detect redundant suggestions. 
Second, the happy moments are still in \hl{the} first person and use descriptive \camr{or potentially de-anonymizing language}. \hl{For example, the activity ``I played ball with our Golden Retriever Rosie'' is labeled and predicted as suggestible, but it should be rephrased as ``play ball with your dog'' before it is presented to a user.}

\section{Related Work}\label{sec:related_work}
Textual analysis for understanding happiness is an emerging topic in the intersection between NLP and psychology fields. Asai et al.~\cite{Asai:2018:LREC} collected 100k happy moments called HappyDB using MTurk. CL-Aff Shared Task 2019 \cite{Jaidka:2019:AffCon} collected another set corpus based on HappyDB, contributing two additional labels on each happy moment: agency and sociality, \camr{which we utilized in this work}. Several studies have applied NLP techniques to the corpora to understand the cause of happiness. Rajendran et al.~\cite{Rajendran:2019:AffCon} have tested various machine learning techniques including NN models for the sociality and the agency classification tasks, and they confirm that NN methods robustly performed better for these tasks. Gupta et al.~\cite{Gupta:2019:Constitutes} have used intensity scores of five emotions (valence, joy, anger, feature, and sadness) in addition to concept vectors for the sociality and agency classification tasks. Their experimental results show that concept vectors significantly contribute to the classification accuracy compared to the emotion\hl{al} intensity scores.

The Recognizing Textual Entailment (RTE) task~\cite{dagan2005pascal} is defined as recognizing whether the meaning of one text can be inferred (i.e., entailed) from the other based on given two text inputs. The task has been intensively studied in NLP~\cite{snli:emnlp2015, Conneau:2017:EMNLP}.
Although the problem formulation of the HER task is similar to the RTE task, we empirically show that the state-of-the-art pre-trained RTE model does not perform well on the HER task, and thus there is little overlap between these two tasks.

\section{Conclusion and Future Work}\label{sec:conclusion}
\camr{Inspired by our initial work on the Jo application~\cite{Jo}, this paper took a first step towards a module that 
automatically suggests activities that will increase the well-being of its user. 
We developed two essential components of generating suggestions.}
The first is the {\it Happiness Entailment Recognition} (HER) component that enables one to determine, based on textual analysis, whether a sustainable suggestion is suitable given a happy moment. 
Our results show that our dual-encoder HER model with the concepts, agency and sociality labels achieve the highest accuracy compared to the baseline methods, supporting the idea that we need to train a dedicated model for the HER task.  
%
For the second component, we developed a sustainable suggestion classifier using machine learning and we have shown that the model can detect sustainable activities in happy moments.

\camr{Future work includes incorporating more sophisticated information about the user in proposing activities,  including the user's history of happy moments, their responses to suggestions, and possibly other contextual information such as location, weather, mood, etc. Collection of more diverse suggestions, whether curated or automatically extracted by expanding on our techniques, will be critical to the success of this system.
}
Further investigation is also required to understand the interactions between agency, sociality, and concept labels with respect to HER, and to explore other psychological features that may be important.


\bibliography{happydb}

\begin{thebibliography}{10}
\providecommand{\url}[1]{#1}
\csname url@samestyle\endcsname
\providecommand{\newblock}{\relax}
\providecommand{\bibinfo}[2]{#2}
\providecommand{\BIBentrySTDinterwordspacing}{\spaceskip=0pt\relax}
\providecommand{\BIBentryALTinterwordstretchfactor}{4}
\providecommand{\BIBentryALTinterwordspacing}{\spaceskip=\fontdimen2\font plus
\BIBentryALTinterwordstretchfactor\fontdimen3\font minus
  \fontdimen4\font\relax}
\providecommand{\BIBforeignlanguage}[2]{{%
\expandafter\ifx\csname l@#1\endcsname\relax
\typeout{** WARNING: IEEEtran.bst: No hyphenation pattern has been}%
\typeout{** loaded for the language `#1'. Using the pattern for}%
\typeout{** the default language instead.}%
\else
\language=\csname l@#1\endcsname
\fi
#2}}
\providecommand{\BIBdecl}{\relax}
\BIBdecl

\bibitem{Duckworth:2005:positive}
A.~L. Duckworth, T.~A. Steen, and M.~E. Seligman, ``Positive psychology in
  clinical practice,'' \emph{Annu. Rev. Clin. Psychol.}, vol.~1, pp. 629--651,
  2005.

\bibitem{Seligman:2012:Flourish}
M.~E. Seligman, \emph{Flourish: A visionary new understanding of happiness and
  well-being}.\hskip 1em plus 0.5em minus 0.4em\relax Simon and Schuster, 2012.

\bibitem{Fredrickson:1998:WhatGood}
B.~L. Fredrickson, ``What good are positive emotions?'' \emph{Review of general
  psychology}, vol.~2, no.~3, pp. 300--319, 1998.

\bibitem{Lyubomirsky:2005:Benefits}
S.~Lyubomirsky, L.~King, and E.~Diener, ``The benefits of frequent positive
  affect: Does happiness lead to success?'' \emph{Psychological bulletin}, vol.
  131, no.~6, p. 803, 2005.

\bibitem{Headey:2010:HappinessAsAChoice}
B.~Headey, R.~Muffels, and G.~G. Wagner, ``Long-running german panel survey
  shows that personal and economic choices, not just genes, matter for
  happiness,'' \emph{Proceedings of the National Academy of Sciences}, 2010.

\bibitem{Lyubomirsky:2013:SimplePositiveActivities}
S.~Lyubomirsky and K.~Layous, ``How do simple positive activities increase
  well-being?'' \emph{Current directions in psychological science}, vol.~22,
  no.~1, pp. 57--62, 2013.

\bibitem{Kahneman:2004:DRM}
D.~Kahneman, A.~B. Krueger, D.~A. Schkade, N.~Schwarz, and A.~A. Stone, ``A
  survey method for characterizing daily life experience: The day
  reconstruction method,'' \emph{Science}, vol. 306, no. 5702, pp. 1776--1780,
  2004.

\bibitem{Larson:2014:ESM}
R.~Larson and M.~Csikszentmihalyi, ``The experience sampling method,'' in
  \emph{Flow and the foundations of positive psychology}.\hskip 1em plus 0.5em
  minus 0.4em\relax Springer, 2014, pp. 21--34.

\bibitem{Shiffman:2008:EMS}
S.~Shiffman, A.~A. Stone, and M.~R. Hufford, ``Ecological momentary
  assessment,'' \emph{Annu. Rev. Clin. Psychol.}, vol.~4, pp. 1--32, 2008.

\bibitem{Diener:2000:SubjectiveWellBeing}
E.~Diener, ``Subjective well-being: The science of happiness and a proposal for
  a national index.'' \emph{American psychologist}, vol.~55, no.~1, p.~34,
  2000.

\bibitem{Jo}
V.~Li, A.~Halevy, A.~Zief-Balteriski, W.-C. Tan, G.~Mihaila, J.~Morales,
  N.~Nuno, H.~Liu, C.~Chen, X.~Ma, S.~Robins, and J.~Johnson, ``Jo: {T}he smart
  journal,'' \emph{arXiv:1907.07861}, 2019.

\bibitem{Asai:2018:LREC}
A.~Asai, S.~Evensen, B.~Golshan, A.~Halevy, V.~Li, A.~Lopatenko, D.~Stepanov,
  Y.~Suhara, W.-C. Tan, and Y.~Xu, ``{HappyDB: A Corpus of 100,000 Crowdsourced
  Happy Moments},'' in \emph{Proc. LREC '18}, 2018.

\bibitem{Jaidka:2019:AffCon}
K.~Jaidka, S.~Mumick, N.~Chhaya, and L.~Ungar, ``The {CL-Aff} happiness shared
  task: Results and key insights,'' in \emph{Proc. AffCon '19}, 2019.

\bibitem{Lyubomirsky:2008:TheHowOfHappiness}
S.~Lyubomirsky, \emph{The how of happiness: A scientific approach to getting
  the life you want}.\hskip 1em plus 0.5em minus 0.4em\relax Penguin, 2008.

\bibitem{albee1959mental}
G.~W. Albee, ``Mental health manpower trends: A report to the staff director,
  jack r. ewalt.'' 1959.

\bibitem{Pennebaker:1986:ExpressiveWriting}
J.~W. Pennebaker and S.~K. Beall, ``Confronting a traumatic event: toward an
  understanding of inhibition and disease.'' \emph{Journal of abnormal
  psychology}, vol.~95, no.~3, p. 274, 1986.

\bibitem{dagan2005pascal}
I.~Dagan, O.~Glickman, and B.~Magnini, ``The pascal recognising textual
  entailment challenge,'' in \emph{Machine Learning Challenges Workshop}, 2005,
  pp. 177--190.

\bibitem{snow2008cheap}
R.~Snow, B.~O'Connor, D.~Jurafsky, and A.~Y. Ng, ``Cheap and fast---but is it
  good?: evaluating non-expert annotations for natural language tasks,'' in
  \emph{Proc. EMNLP '08}.\hskip 1em plus 0.5em minus 0.4em\relax Association
  for Computational Linguistics, 2008, pp. 254--263.

\bibitem{Conneau:2017:EMNLP}
A.~Conneau, D.~Kiela, H.~Schwenk, L.~Barrault, and A.~Bordes, ``Supervised
  learning of universal sentence representations from natural language
  inference data,'' in \emph{Proc. EMNLP '17}, 2017, pp. 670--680.

\bibitem{Gluck:2011:LearningAndMemory}
M.~A. Gluck, C.~E. Myers, and E.~Mercado, \emph{Learning and Memory: From Brain
  To Behavior (3rd edition)}.\hskip 1em plus 0.5em minus 0.4em\relax Worth
  Publishers, 2016.

\bibitem{needtobelong}
R.~F. Baumeister and M.~R. Leary, ``The need to belong: desire for
  interpersonal attachments as a fundamental human motivation.''
  \emph{Psychological bulletin}, vol. 117, no.~3, p. 497, 1995.

\bibitem{cantor200312}
N.~Cantor and C.~A. Sanderson, ``12 life task participation and well-being: The
  importance of taking part in daily life,'' \emph{Well-being: Foundations of
  hedonic psychology}, p. 230, 2003.

\bibitem{Parikh:2016:AllenNLPTE}
A.~P. Parikh, O.~T{\"a}ckstr{\"o}m, D.~Das, and J.~Uszkoreit, ``A decomposable
  attention model for natural language inference,'' in \emph{Proc. EMNLP '16},
  2016, pp. 2249--2255.

\bibitem{snli:emnlp2015}
S.~R. Bowman, G.~Angeli, C.~Potts, and C.~D. Manning, ``A large annotated
  corpus for learning natural language inference,'' in \emph{Proc. EMNLP '15},
  2015.

\bibitem{Bergstra:2012:RandomSearch}
J.~Bergstra and Y.~Bengio, ``Random search for hyper-parameter optimization,''
  \emph{Journal of Machine Learning Research}, vol.~13, no. Feb, pp. 281--305,
  2012.

\bibitem{Stern:2017:AllenNLPConstituencyParser}
M.~Stern, J.~Andreas, and D.~Klein, ``A minimal span-based neural constituency
  parser,'' in \emph{Proc. ACL '17}, 2017, pp. 818--827.

\bibitem{Lin:2017:SelfAttention}
Z.~Lin, M.~Feng, C.~N.~d. Santos, M.~Yu, B.~Xiang, B.~Zhou, and Y.~Bengio, ``A
  structured self-attentive sentence embedding,'' in \emph{Proc. ICLR '17},
  2017.

\bibitem{Rajendran:2019:AffCon}
M.~A.-M. Arun~Rajendran, Chiyu~Zhang, ``{Happy Together}: {L}earning and
  understanding appraisal from natural language,'' in \emph{Proc. AffCon '19},
  2019.

\bibitem{Gupta:2019:Constitutes}
R.~K. Gupta, P.~Bhattacharya, and Y.~Yang, ``What constitutes happiness?
  predicting and characterizing the ingredients of happiness using emotion
  intensity analysis,'' in \emph{Proc. AffCon '19}, 2019.

\end{thebibliography}
\bibliographystyle{IEEEtran}

\end{document}